\documentclass[10pt,twocolumn,a4paper]{article}
\usepackage{float}
\usepackage[utf8]{inputenc}
\usepackage[T1]{fontenc}
\usepackage{amsmath,amssymb,amsfonts}
\usepackage{graphicx}
\usepackage{geometry}
\usepackage{hyperref}
\usepackage{cite}
\usepackage{setspace}
\usepackage{caption}
\usepackage{balance}
\title{The geometry of forgetting and plasticity in continual learning:   A boundary-anchored three-protocol analysis framework}
\author{
\begin{tabular}{@{}p{0.95\textwidth}@{}}
Songhan Li$^{a,1}$ \quad
Jiaxin Qi$^{c, 1}$ \quad
Chunguo Wu$^{a, b, 2}$ \quad
Bo Yang$^{a, b, c, 2}$\\[0.4em]
{\small $^{a}$College of Computer Science and Technology,
Jilin University,
Changchun, China}\\
{\small $^{b}$MOE Key Laboratory of Symbolic Computation and Knowledge Engineering,
Jilin University,
Changchun, China}\\
{\small $^{c}$College of Software, Jilin University,
Changchun, China}
\end{tabular}
}
\date{}

\begin{document}

\maketitle

\noindent{\bfseries Deep neural networks in continual learning face a fundamental dilemma: sequential training unavoidably trades off the retention of prior knowledge against the ability to acquire new knowledge. Conventional evaluation relies on coarse scalar metrics that obscure the geometric reorganization of internal representations—the way feature representations and classifier weights rearrange in their geometric structure across tasks. Here, we establish a boundary-anchored three-protocol analysis framework—encompassing independent, continual, and joint training—to evaluate continual learning algorithms against two interpretable extremes: independent training defines an idealized neural-collapse bound free from cross-task interference; joint training establishes an empirical structural ceiling attainable when all data are available simultaneously. Anchored against these two boundaries, we introduce a multidimensional geometric metric suite adapted from the neural collapse criteria ($\text{NC}_1$–$\text{NC}_4$), with $\text{NC}_2$ decomposed into Equiangularity (EA) and Equinormality (EN). Systematic evaluation of Classical Backpropagation(BP), Backpropagation with L2-norm penalty (L2), Continual Backpropagation (CBP), and Elastic Weight Consolidation (EWC) reveals that continual learning universally perturbs representational geometry, most notably through the progressive decoupling of classifier weights and class means captured by $\text{NC}_3$. The EWC pursues stability at the cost of explosive growth of $\text{NC}_1$, exceeding even the joint-training ceiling of within-class compactness. The EA–EN decomposition further discovers a critical asymmetry: \textbf{angular arrangements} retain substantial structural inertia, while \textbf{feature norms} are acutely vulnerable to sequential interference—the perturbation of previously learned representations caused by subsequent task updates. By transforming continual learning evaluation from scalar performance tracking into a rigorous, mechanism-level structural diagnosis, this work furnishes a principled geometric foundation for the mechanistic design of future continual learners.\par}

\vspace{\baselineskip}

\section{Introduction}

\footnotetext[1]{J.Q. contributed equally to this work with S.L.}
\footnotetext[2]{To whom correspondence may be addressed.\hfill\\
E-mail: \texttt{wucg@jlu.edu.cn}; \texttt{ybo@jlu.edu.cn}.}

While deep neural networks have achieved remarkable success across computer vision and beyond\cite{ref1,ref2}, their deployment in continual learning scenarios remains hindered by a fundamental tension between two core properties: catastrophic forgetting and plasticity loss \cite{ref3,ref4,ref5,ref6}. On one hand, models trained sequentially on multiple tasks rapidly degrade on previously learned mappings, reflecting poor stability in preserving old task knowledge. On the other hand, excessive stability often impairs plasticity, namely the model’s ability to acquire new task knowledge continuously. Over the past decade, a proliferation of algorithms\cite{ref3,ref4}—from regularization-based methods such as EWC\cite{ref5} and CBP\cite{ref7} to architectural and replay-based approaches—have sought to reconcile stability with plasticity\cite{ref8,ref9}. Yet the community predominantly evaluates these methods through scalar downstream
metrics\cite{ref3,ref24}, such as final average accuracy\cite{ref22}, forgetting rate\cite{ref23}, and backward transfer\cite{ref22}. These measures effectively compress algorithmic behavior into a single metric, summarizing whether a model performs well while providing limited insight into the underlying representational mechanisms.

To understand the underlying representational mechanisms, we propose a \textbf{boundary-anchored tri-protocol analysis framework} that re-anchors continual learning evaluation in the internal geometry of deep representations rather than in external task accuracy alone. Our starting point is the neural collapse (NC) phenomenon: as a deep classifier approaches the terminal phase of training, features of the same class collapse toward their class mean, the class means converge to a simplex equiangular tight frame (ETF), and the classifier weights align with the corresponding class means\cite{ref10,ref11,ref14,ref18}. This terminal geometry offers a task-level description of how well a network has organized its representation for classification. The framework converts this description into two reference protocols. In independent training, each task is learned in isolation, so the network can reach a clean, task-specific NC geometry without cross-task interference. In joint training, all tasks are learned concurrently from the full data, so the network can organize a shared representation without sequential forgetting; its geometry serves as the empirical structural ceiling---the best-case geometry that the same architecture actually attains when data are available simultaneously. The ceiling is empirical rather than a mathematical guarantee: it is the geometry realized by the jointly trained model, and it provides the upper reference against which continual learners are compared. Every evaluated continual learning algorithm then falls along an intermediate trajectory between these two anchors, and its deviation from either anchor quantifies the geometric cost incurred by sequential training. We summarize this cost along two complementary symptoms: clustering rigidity, which concerns how tightly within-class features remain clustered around their class means, and angular alignment, which concerns the angular arrangement of class means and their alignment with the corresponding classifier weights. Both symptoms become quantitatively measurable within the framework .

Operating within this tri-protocol analysis framework, we investigate continual learning from the perspective of the internal geometry of deep representations, thereby formulating a suite of diagnostic probes tailored for continual learning evaluation. On the representational perspective, we adapt classic Neural Collapse metrics—including $\text{NC}_1$ (class collapse), $\text{NC}_2$ (convergence to a simplex Equiangular Tight Frame), $\text{NC}_3$ (classifier–feature alignment), and $\text{NC}_4$ (nearest-center-classifier agreement) \cite{ref10}—to the sequential learning setting. Given that classical $\text{NC}_2$ entangles two geometrically distinct properties\cite{ref11,ref12}, we further disentangle the original $\text{NC}_2$ into two orthogonal geometric dimensions: Equiangularity (EA) ,for inter-class angular regularity and Equinormality (EN) ,for feature norm regularity. Each probe characterizes a distinct geometric structural signature: $\text{NC}_1$ quantifies within-class compaction, $\text{NC}_3$ tracks classifier–feature alignment consistency, and EA and EN collectively dissect the geometric organization of class means. Operating jointly within the bounds established by the independent and joint training protocols, these diagnostic probes convert the proposed framework from a conceptual formulation into a fully quantitative tool for mechanism-level analysis.

Leveraging this unified set of geometric probes, our tri-protocol analysis reveals explicit geometric characteristics of continual learning dynamics. $\text{NC}_3$ indicates a universal trend across all evaluated methods: classifier weights and class means gradually diverge from each other, with the degree of separation consistently growing as the task number increases (Figure 2). Notably, CBP and L2 substantially alleviate such divergence, demonstrating their capability to better preserve classifier–feature alignment throughout sequential training. t-SNE visualization of $\text{NC}_3$\cite{ref13} further intuitively verifies the gradual detachment of classifier weights from their task-evolved class means of representational features. In-depth analysis of the disentangled EA and EN dimensions further uncovers how continual learning asymmetrically distorts feature geometry. Specifically, EA undergoes only minor drift across sequential tasks (Figure 4), suggesting that the angular configuration of class means retains strong structural inertia against sequential interference. In sharp contrast, EN is highly susceptible to continual learning perturbations: different algorithms rescale feature magnitudes differently and disrupt the inherent norm balance. Since angular and norm configurations constitute partially independent geometric dimensions, simply aggregating them into a single $\text{NC}_2$ scalar inevitably masks this crucial asymmetry in representational stability. 

Collectively, our boundary-anchored tri-protocol framework transforms conventional continual learning evaluation from an accuracy-centric black-box assessment into a geometry-aware, interpretable analysis paradigm. By characterizing algorithmic behavioral patterns across the independent–continual–joint training continuum via multi-dimensional Neural Collapse metrics and their orthogonal extensions, this work constructs a systematic diagnostic framework for revealing how model stability is achieved at the cost of representational distortion. Furthermore, it provides a principled mechanism-level foundation for the mechanistic design and performance optimization of future continual learning models.

\section{Results}

\subsection{Forgetting Suppression at the Cost of Within-Class Compactness ($\text{NC}_1$)}
Within our boundary-anchored tri-protocol analysis framework, the independent training baseline defines the ideal lower boundary for $\text{NC}_1$, corresponding to optimal within-class compactness free from cross-task interference; while joint training establishes the empirical structural upper boundary. As illustrated in Figure 1, the continual learning trajectories of most evaluated algorithms consistently lie between these two structural boundaries. However, Figure 1(D) demonstrates a distinct geometric anomaly observed in EWC. Although EWC effectively suppresses catastrophic forgetting throughout sequential task learning\cite{ref5}, it yields persistent and substantial $\text{NC}_1$ elevation that rapidly surpasses the joint-training upper boundary as the task number increases. In contrast, alternative methods including BP (Figure 1(A)), CBP (Figure 1(B)), and L2 (Figure 1(C)) exhibit far more stable $\text{NC}_1$ within-class compactness metrics than EWC across sequential training with up to 30 tasks. Their $\text{NC}_1$ values fluctuate consistently around the reference level derived from independent training; notably, CBP (Figure 1(B)) and L2 (Figure 1(C)) show markedly milder fluctuations compared to BP (Figure 1(A)). This empirical observation validates a core mechanistic finding: EWC realizes stable memory retention by rigidly restricting parameter update magnitudes, which substantially compromises the model’s capacity to maintain compact within-class feature representations for sequentially acquired tasks.

\begin{figure}[t]
    \centering
   \includegraphics[width=\columnwidth]{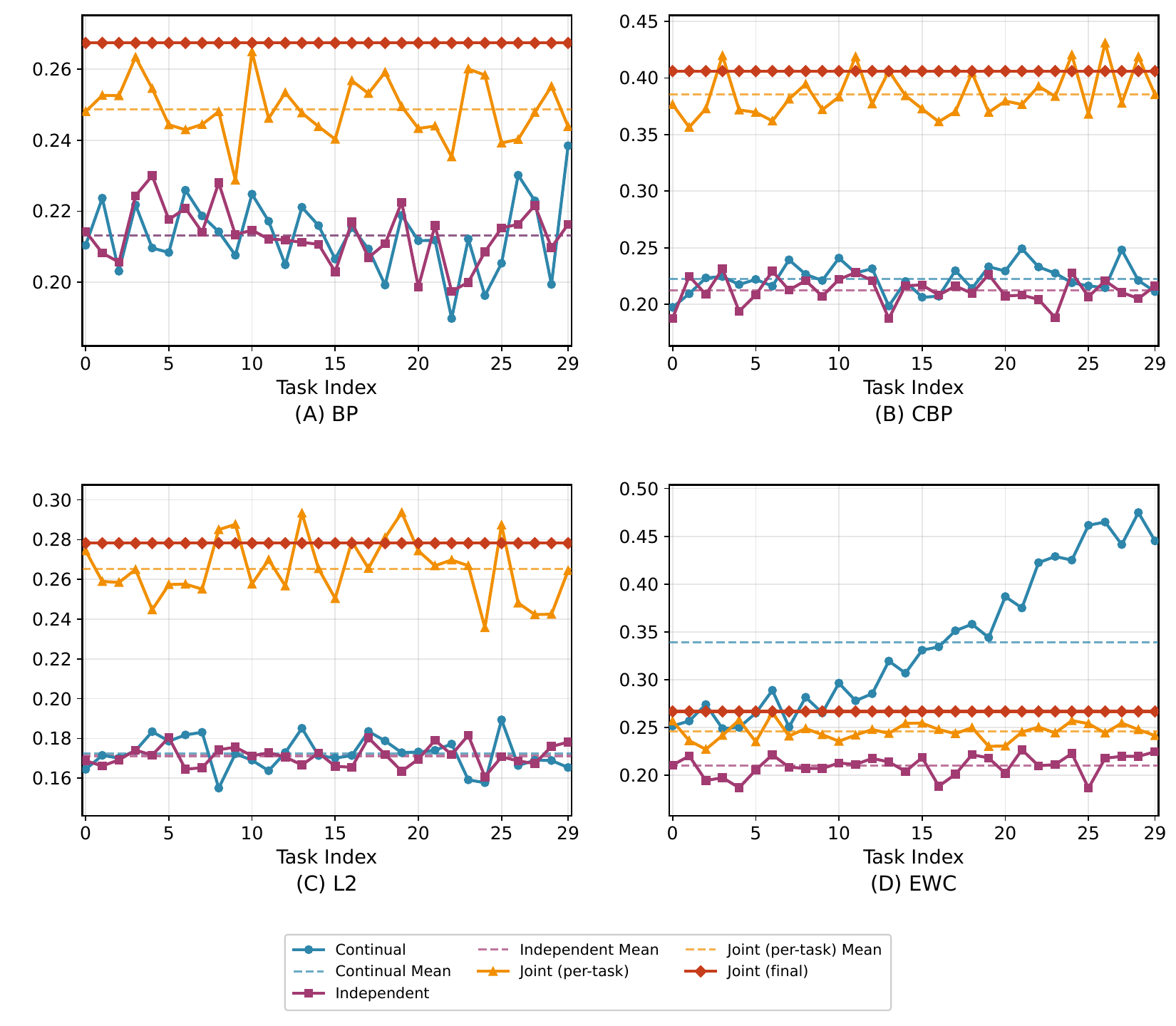}
    \caption{Evolution of $\boldsymbol{\text{NC}_1}$ with task index under different learning algorithms. The horizontal axis denotes task index, and the vertical axis denotes neural collapse metric $\text{NC}_1$. The blue and purple curves correspond to the continual learning and independent training, respectively. The orange curve is for per‑task joint training results, obtained by testing trained tasks on the network derived from joint training. The red dashed line corresponds to the final joint training baseline. The horizontal dashed lines of matching colors represent the average value of the corresponding curves. Subplots (A)–(D) correspond to BP, CBP, L2, and EWC algorithms, respectively.}
  \label{fig:forward_transfer}
\end{figure}

\subsection{Universal degradation of classifier--feature duality ($\text{NC}_3$): a shared direction, an algorithm-dependent magnitude}

Within our boundary-anchored tri-protocol analysis framework, we observe a universal degradation of the classifier--feature duality ($\text{NC}_3$)---the structural coupling between classifier weights, which define the decision boundaries of each task, and task-evolved class means, which constitute the core of the internal feature distributions. As illustrated by the $\text{NC}_3$ trajectories in Figure 2, this misalignment drifts upward across all four evaluated algorithms: BP, EWC, L2, and CBP. Class means are continuously reshaped under sequential cross-task interference and gradually decouple from the classifier weights with which they were originally aligned. This shared direction is not incidental: none of the four learners contains a term that explicitly re-anchors the classifier head to the displaced class means, so every update rule inherits the same drift, and $\text{NC}_3$ registers it in every case.

The direction of the drift is universal; its magnitude is not, and the magnitude is not a matter of update strength. This decoupling is a direct geometric consequence of sequential interference, and the learners diverge on a single question: whether the update rule leaves the classifier subspace free to turn with the feature space. When a new task arrives, the shared embedding is reconstructed and the class means of earlier tasks are displaced to new locations. Each classifier weight $w_k$ is a direction in weight space that defines a decision boundary, and $\text{NC}_3$ is the cosine between that direction and the displaced mean: the metric is sensitive to relative orientation and blind to overall scale. An update rule can deny the head that freedom in two opposite ways---by driving it exclusively with the incoming task, or by freezing it in a configuration inherited from an earlier task---yet the geometric outcome is identical: an ever-widening spatial gap between where a class actually resides in the feature space and where its classifier weight points.


Both ways of denying the head that freedom appear among our four learners, and so do two ways of granting it. BP updates the extractor and the head with a gradient computed only on the new task, so the head follows the incoming classes while the old means are left behind. EWC instead anchors parameters to their previous optimum through Fisher weights, and because that anchor is directionally selective, the head is held in an orientation the features have already abandoned. L2 shows that a parameter penalty need not produce this outcome: the $\text{L}_2$ penalty is isotropic, shrinking every direction with equal proportional force and imposing no preferred orientation, so the head remains free to co-adapt with the drifting means even though the update itself stays small. CBP attains the same freedom from the opposite direction, replacing low-utility parameters so that the parameter space never rigidifies and the classifier can keep tracking the evolving class means.


Figure 2 shows the consequence at the level of the metric itself. All four trajectories drift upward, yet they separate into two clearly different magnitudes of degradation. BP climbs most steeply and converges toward the joint-training level, indicating that nearly the whole of the duality is lost as tasks accumulate; EWC rises more slowly but continuously, and the gap between the two trajectories widens rather than closes with task index. The trajectories of L2 and CBP, by contrast, saturate: they rise over the first few tasks and then settle onto a plateau that neither drifts nor widens as the sequence continues, and that plateau stays far below the joint-training level. This early saturation is the visual signature of a mechanism that permits the classifier head to track, rather than resist, the evolving class means. The per-class extremes (triangles) move in step with the mean in all four panels, confirming that the separation between learners reflects the class configuration as a whole rather than a few anomalous classes: the entire envelope of BP is displaced upward, whereas the envelopes of L2 and CBP remain within a band far below the joint-training level.


\begin{figure}[t]
    \centering
   \includegraphics[width=\columnwidth]{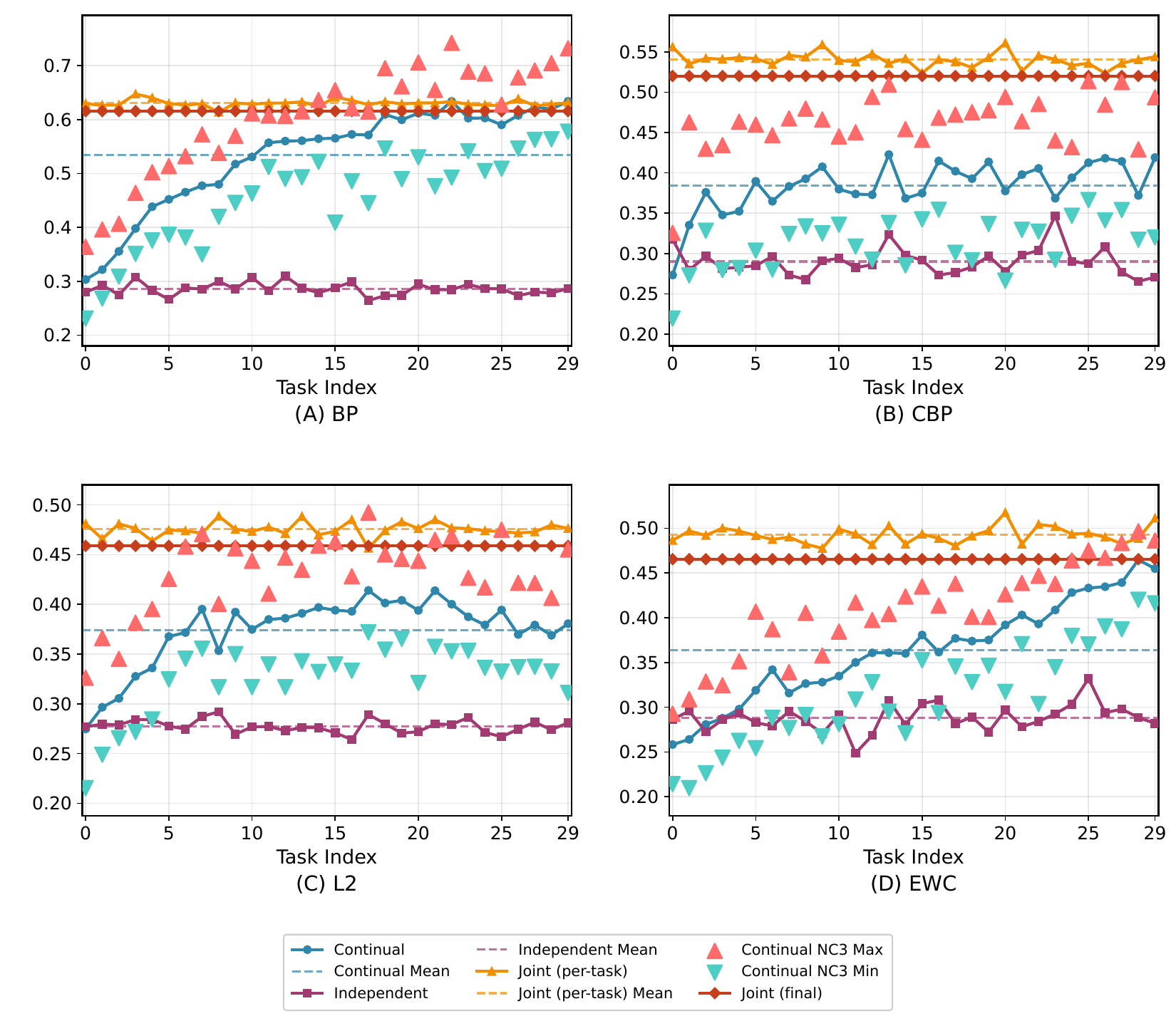}
    \caption{Evolution of ${\text{NC}_3}$ with task index under different continual learning algorithms. The horizontal axis denotes task index, and the vertical axis denotes the neural collapse metric $\text{NC}_3$ that characterizes classifier‑feature duality. The blue solid line represents the mean $\text{NC}_3$ under the continual learning protocol, while the red upward triangles and cyan downward triangles correspond to the maximum and minimum $\text{NC}_3$ values among ten classes under the continual learning protocol, which are used to characterize metric variance and avoid result bias caused by abnormal offsets of individual classes. The purple solid line shows results of the independent training protocol, the orange solid line denotes per‑task joint training results obtained by testing trained tasks on the network derived from joint training, and the red solid line represents the final joint training baseline.The horizontal dashed lines of matching colors represent the average value of the corresponding curve }
  \label{fig:forward_transfer}
\end{figure}

Figure 3 renders this gap visually explicit. The t-SNE visualizations of class-center and classifier-weight embeddings show that from the initial task (Figure 3(A)) to the final task (Figure 3(D)), the distance between class centers (circles) and their corresponding classifier weights (squares) grows monotonically. This universal geometric divergence is precisely what the rising $\text{NC}_3$ scalar quantifies: a physical decoupling in the feature space where classifier weights and class means increasingly fail to align.
\begin{figure}[t]
    \centering
    
   \includegraphics[width=\columnwidth]{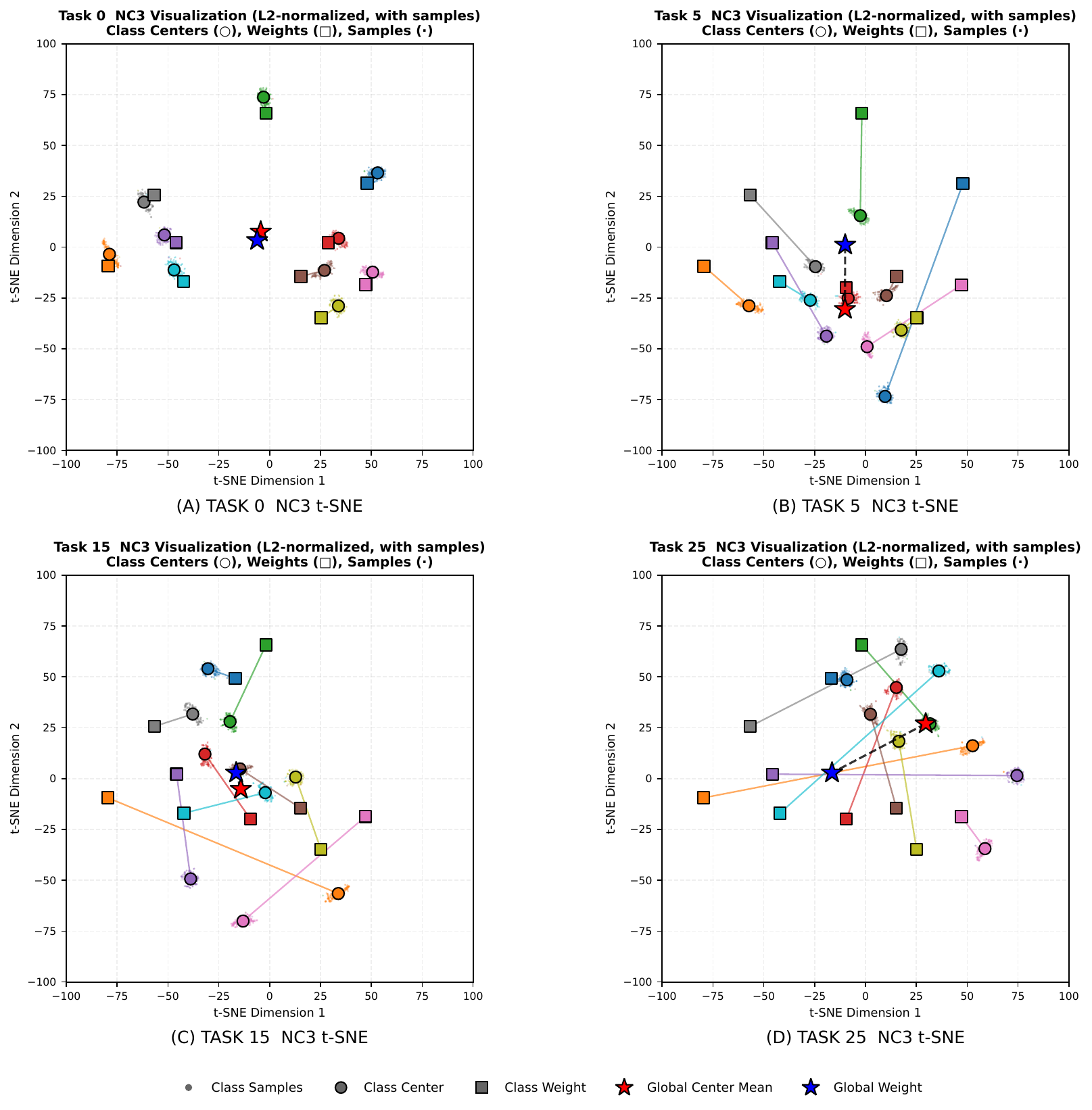}
    \caption{t‑SNE‑based feature space visualization of ${\text{NC}_3}$ (taking 25 continual learning tasks as an example). The horizontal and vertical axes represent the first and second feature dimensions after t‑SNE dimensionality reduction, with all features $\text{L}_2$‑normalized. Colorful scatter points denote samples of each class, circular markers represent class centers, square markers correspond to classifier weights of respective classes, and straight lines connect class centers and classifier weights within the same class. Subplots (A)–(D) sequentially show visualization results of the 0th, 5th, 15th, and 25th tasks, intuitively demonstrating the geometric evolution of gradual separation between class centers and classifier weights in the feature space as tasks accumulate.}
  \label{fig:forward_transfer}
\end{figure}


\subsection{Distinct geometric costs in $\text{NC}_2$ decomposition}
Finally, we demonstrate the necessity of decomposing the original aggregated $\text{NC}_2$ metric into two orthogonal components: Equiangularity (EA) and Equinormality (EN)\cite{ref11,ref12}. Evaluated within our tri-protocol framework, as illustrated in Figure 4, these metrics expose a critical asymmetry in representational stability. The angular arrangement of the class means, captured by EA, exhibits substantial structural inertia, showing only mild drift across sequential tasks and remaining relatively stable. In contrast, EN is much more vulnerable to continual-learning interference. Algorithms stretch or compress feature scales differentially across tasks, severely disrupting the norm balance of the representations. Because angle and magnitude structures behave as partially independent geometric dimensions responding differently to sequential pressure, aggregating them into a single scalar $\text{NC}_2$ obscures the true mechanism of structural degradation. Observing them separately provides a significantly clearer and more precise diagnosis of how different algorithms perturb the geometry of internal features.

\begin{figure}[t]
    \centering
   \includegraphics[width=\columnwidth]{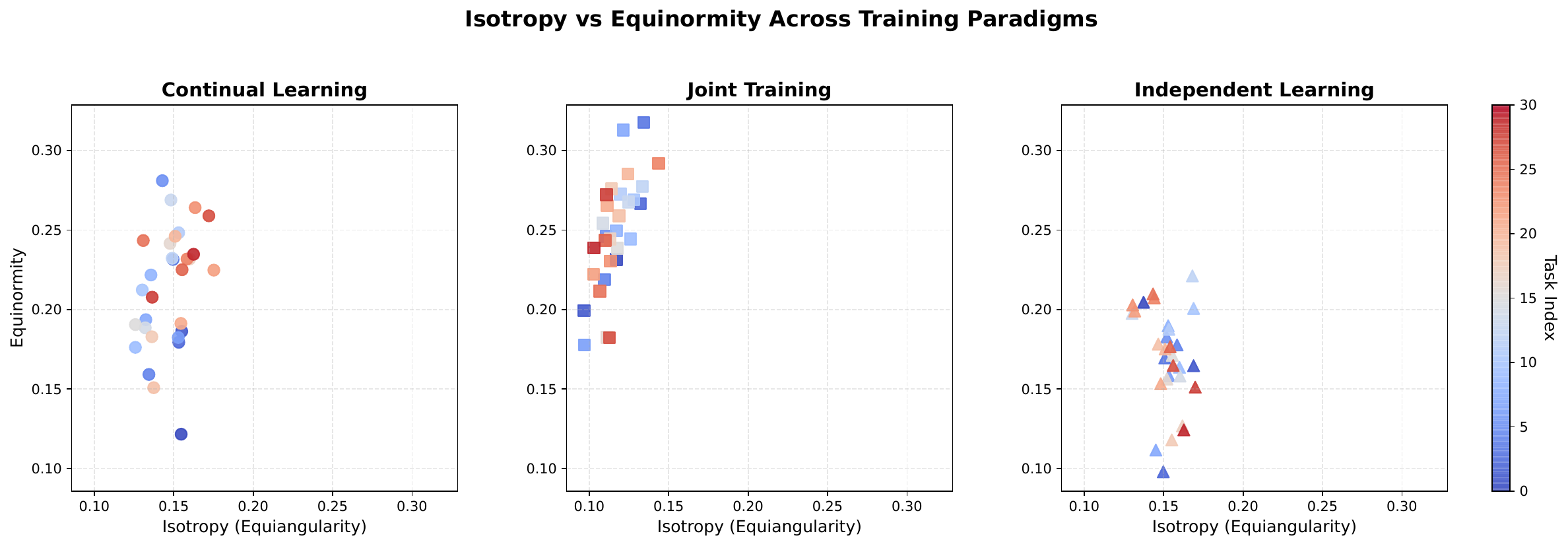}
    \caption{Two‑dimensional scatter distribution of equiangularity (EA) and equinormality (EN) under three training protocols. The three subplots correspond to continual learning, joint training, and independent learning protocols from left to right. The horizontal axis denotes the isotropy (EA) metric characterizing the angular structure between classes, while the vertical axis represents the equinormality (EN) metric characterizing feature norm balance. Scatter point colors are mapped to task index via the right‑side color bar (blue for early tasks, red for late tasks). Circular, square, and triangular markers correspond to sample points of continual learning, joint training, and independent learning protocols, respectively.}
  \label{fig:forward_transfer}
\end{figure}

%
%

\subsection{Cross-architecture validation on ResNet: portable duality and protocol-dependent compactness}
\label{sec:resnet}

The preceding analyses establish the geometric signatures of sequential learning on a fully connected network. A natural question is whether these signatures are specific to that architecture or reflect a more general property of sequential learning. To answer this question, we carry out the tri-protocol analysis on ResNet, a convolutional backbone with residual connections and batch normalization, using TinyImageNet as a natural-image benchmark. The task sequence consists of 30 five-class tasks drawn from 150 TinyImageNet classes, and the identity of the task is known at both training and evaluation. The network comprises four residual stages, and all learners share the same class order and evaluation protocol. We compare BP, L2, CBP, and PredKD. The first three instantiate the same algorithm families as in the primary benchmark, while PredKD introduces a regularization strategy that acts on the predictive distribution rather than on the parameters.

PredKD is a regularization-based learner that transfers the predictive behavior of the previous task's model to the current one\cite{ref19}. At each task boundary, the model trained on the preceding task is frozen as a teacher, and the learner minimizes a temperature-scaled divergence between the teacher's and the student's logits in addition to the standard cross-entropy objective\cite{ref20,ref21}. This constraint differs from the EWC in both locus and effect. EWC penalizes displacement in parameter space according to Fisher importance, whereas PredKD penalizes displacement in function space, that is, changes in the network's response to inputs from earlier tasks. By preserving the old predictive relations without rigidly anchoring individual weights, PredKD leaves the feature extractor free to adapt to the new task while discouraging the classifier--feature configuration from drifting away from its previous state. This mechanism provides a natural account of why PredKD retains forgetting resistance while preserving favorable classifier--feature duality and within-class compactness: the distillation term protects the old decision structure, and the unconstrained parameter mobility allows the shared representation to accommodate new classes without collapsing the class means of earlier tasks.

\begin{figure}[t]
    \centering
    \includegraphics[width=\columnwidth]{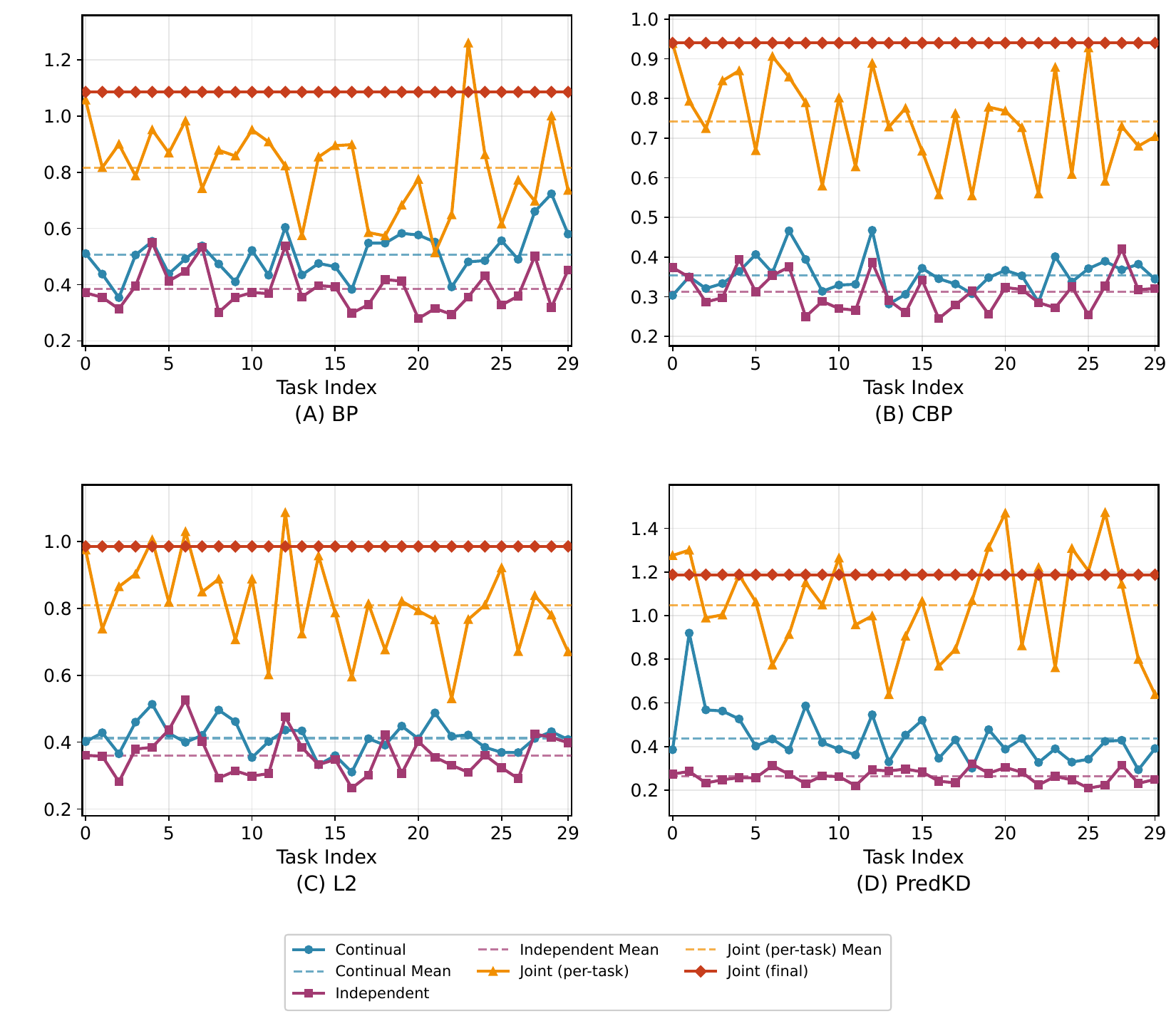}
    \caption{Evolution of $\text{NC}_1$ with task index under different learning algorithms on ResNet-18/TinyImageNet. The horizontal axis denotes task index (30 five-class tasks drawn from 150 TinyImageNet classes), and the vertical axis denotes $\text{NC}_1$. The blue and purple curves correspond to continual and independent training, respectively. The orange curve is the per-task joint-training result, obtained by evaluating each trained task on the jointly trained network. The red horizontal line is the final joint-training baseline. The horizontal dashed lines of matching colors represent the average value of the corresponding curves. Subplots (A)--(D) correspond to BP, L2, CBP, and PredKD.}
    \label{fig:resnet_nc1}
\end{figure}

Figure~\ref{fig:resnet_nc1} reports the evolution of $\text{NC}_1$ under the three protocols. The overall picture is consistent with the primary benchmark: for all four learners the continual trajectory stays between the independent and joint reference levels, and BP, L2, and CBP reproduce their familiar patterns, with BP ranging widely between the two references and the two regularized learners staying flatter near the independent level. The PredKD panel departs from this pattern in one place. Its $\text{NC}_1$ spikes at the second task and then relaxes and stabilizes for the rest of the sequence. The spike is a consequence of the distillation constraint. The teacher for task 1 is the task-0 model, and the teacher-student loss is evaluated on inputs from the incoming task, so the student is briefly required to reproduce the old model's responses to a different class set while the shared representation has not yet adapted; the class means are stretched away from their centers, and $\text{NC}_1$ rises accordingly. Once the feature extractor accommodates the new classes, the distillation term and the classification objective become compatible, the class means re-tighten, and the metric returns to a stable level below the joint reference. The transient is confined to the start of the sequence because it is there that the teacher has seen the fewest tasks and the representation has the least room to absorb a new class set.

\begin{figure}[t]
    \centering
    \includegraphics[width=\columnwidth]{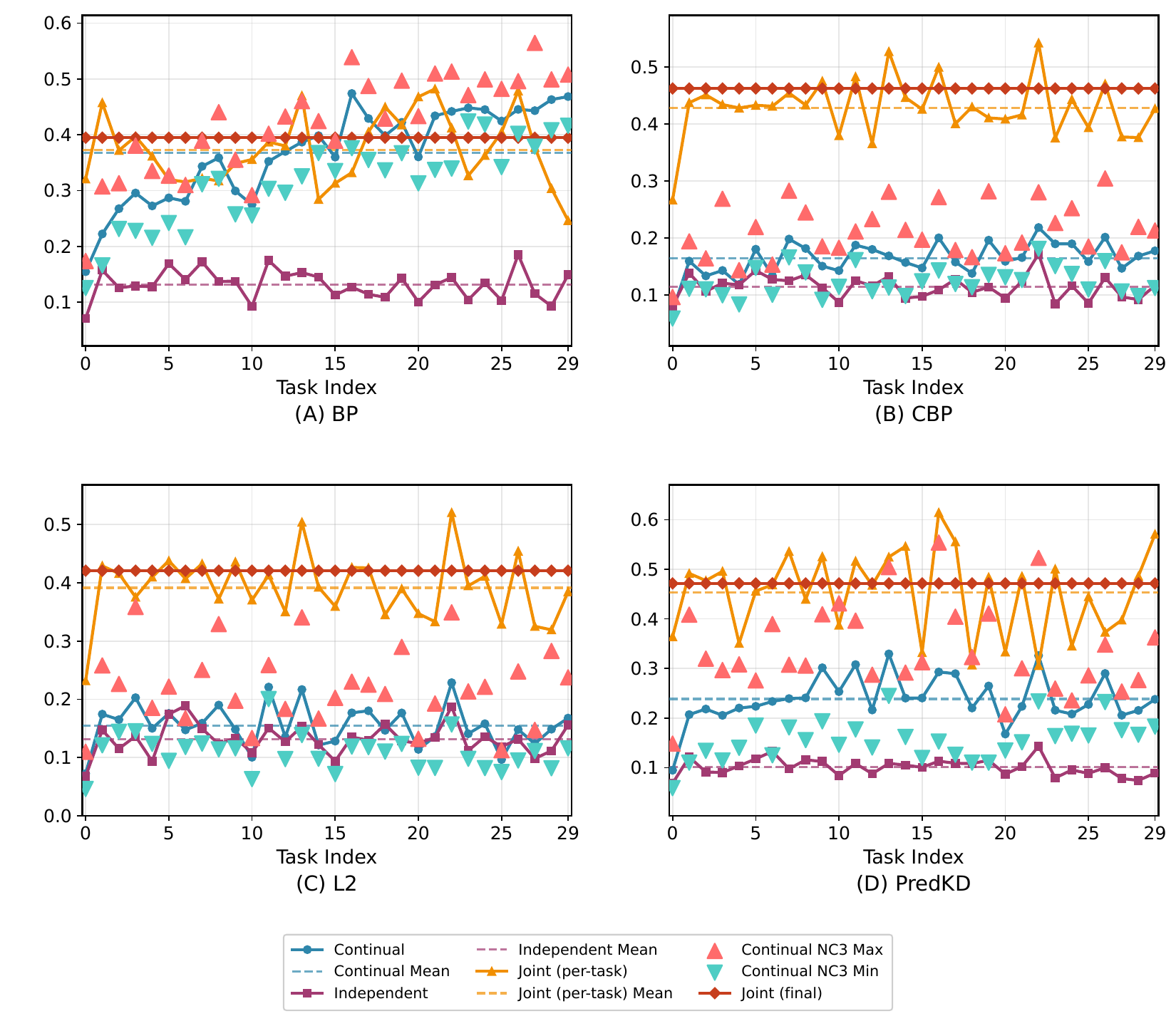}
    \caption{Evolution of $\text{NC}_3$ with task index under different learning algorithms on ResNet-18/TinyImageNet. The blue solid line represents the mean $\text{NC}_3$ under continual learning, while the red upward and cyan downward triangles correspond to the maximum and minimum $\text{NC}_3$ values among the five classes. The purple solid line shows independent training, the orange solid line denotes per-task joint training, and the red solid line represents the final joint-training baseline. The horizontal dashed lines of matching colors represent the average value of the corresponding curves. Subplots (A)--(D) correspond to BP, L2, CBP, and PredKD.}
    \label{fig:resnet_nc3}
\end{figure}

Figure~\ref{fig:resnet_nc3} reports the corresponding $\text{NC}_3$ trajectories. The three-protocol bracketing is again reproduced: independent training yields the lowest $\text{NC}_3$, joint training the highest, and the continual trajectory lies between them.BP exhibits a pronounced upward drift, with $\text{NC}_3$ rising steadily as tasks accumulate, whereas L2 and CBP remain close to their independent references throughout the sequence. PredKD also suppresses the drift: its $\text{NC}_3$ stays low and stable, remaining well below the BP trajectory. The contrast between BP and the three regularized or plasticity-preserving learners shows that classifier--feature decoupling is not an inevitable consequence of sequential training on a convolutional backbone; it is controlled by the stability mechanism. In particular, PredKD's functional constraint preserves the old predictive structure and thereby keeps the classifier weights aligned with the evolving class means, while CBP's replacement mechanism preserves parametric mobility and prevents the feature geometry from rigidifying.

Taken together, the ResNet results reproduce the central signatures of the tri-protocol framework on a convolutional architecture and a natural-image task sequence. The compactness axis preserves the independent--continual--joint ordering and the qualitative separation between learners; the duality axis preserves both the ordering and the contrast between drift and stability. PredKD adds a further point to this picture: a learner that constrains the predictive function rather than the parameters can retain forgetting resistance while maintaining favorable classifier--feature duality and within-class compactness. This is consistent with the framework's central claim that stability and plasticity leave distinct geometric signatures, and it shows that these signatures are not artifacts of a particular network family.Full architectural and optimization details are provided in SI Appendix S2.2.

\section{Discussion}

The central contribution of this work is a boundary-anchored evaluation scaffold that relocates the assessment of continual learning from the exterior (task accuracy) to the interior (representation geometry)\cite{ref3}. By measuring neural collapse criteria against the dual references of independent and joint training, we convert a black-box scalar dilemma into a multi-axis structural read-out. Our results show that this internal view is not merely supplementary: two algorithms can exhibit indistinguishable average accuracies while diverging sharply in their $\text{NC}_1$ or $\text{NC}_3$ profiles. Only by inspecting these geometric coordinates can one determine whether poor performance stems from representational dispersion (high $\text{NC}_1$) or classifier--feature misalignment (high $\text{NC}_3$).

Across all benchmarks and algorithms, $\text{NC}_3$ exhibits the most consistent and pronounced degradation under continual training. This universal behavior implicates the decoupling of classifier weights from evolving class means as an intrinsic geometric cost of sequential learning, rather than an artifact of any particular regularization strategy. Mechanistically, the shared classifier head is optimized for new tasks via backpropagation, whereas the feature means of old classes drift as the feature extractor is updated. Because the classifier weights are not explicitly pushed to track these drifting means, the duality between features and decision boundaries systematically erodes. $\text{NC}_3$ therefore serves as a direct, low-dimensional probe of this structural pathology.

EWC and CBP instantiate opposing responses to the forgetting–plasticity tension\cite{ref5,ref7}, and our decomposition reveals that these responses produce orthogonal geometric deformations. EWC enforces parameter-level consolidation via Fisher-weighted penalties, which protects old knowledge. The accompanying geometric signature is a sharp rise in $\text{NC}_1$: features are prevented from collapsing tightly around their means, producing a diffuse representational geometry that accommodates both old and new tasks at the cost of compactness. CBP, by contrast, continually replaces dormant units (i.e., underutilized neurons), which preserves plasticity and significantly attenuates $\text{NC}_3$ degradation. These distinct signatures suggest that future hybrid strategies should monitor multiple geometric axes—$\text{NC}_1$, $\text{NC}_3$, EN, and EA—rather than optimizing a single accuracy–forgetting scalar\cite{ref3}.

A particularly constructive insight arises from the $\text{NC}_2$ decomposition. Joint training establishes a near-ideal equiangularity (EA) ceiling, confirming that maximally regular class-mean angular arrangements are realizable when data are co-available. Remarkably, continual learning induces only mild EA drift relative to this ceiling; the angular regularity among class means exhibits substantial structural inertia under sequential training. In other words, current continual learners do not inherently destroy the equiangular configuration—they simply fail to actively enforce it. This asymmetry, where equinormality is strongly disrupted but equiangularity is largely preserved, points to a concrete and under-explored intervention: explicit geometric constraints that promote or preserve equiangularity—such as angular regularization on class means, feature whitening toward a Simplex ETF\cite{ref11,ref12,ref18}, or architectural biases toward orthogonal representations—could push continual-learning geometry closer to the joint-training ceiling without necessarily interfering with the parameter-level plasticity required for new-task acquisition. 

Broadly speaking, insights can be gained not only from applying this framework to continual-learning algorithms, but also from its observational strategy: treating coarse scalar metrics as surface signals and asking what geometric reorganization lies beneath them. By shifting attention from performance to the geometry that produces it, this perspective offers a new way to understand continual learning and opens a route toward tackling the forgetting--plasticity problem at the level of representation structure.

\section{Setting and Methods}

\subsection{Validation of experimental design against established algorithmic profiles}

Before deploying the geometric diagnostic suite, we first verify that our experimental pipeline recovers the established behavioral signature of EWC. Figure 5 presents the task-by-task test accuracy matrix. After the full task sequence, EWC exhibits the least decay in early-task accuracy (lower-triangular region), directly corroborating its documented anti-forgetting property\cite{ref5}. This external check confirms that our training protocol, network architecture, and evaluation window produce algorithmic footprints consistent with the literature, ensuring that subsequent geometric deviations reflect genuine representational mechanisms rather than experimental artifacts.
\begin{figure}[t]
    \centering
    \includegraphics[width=\columnwidth]{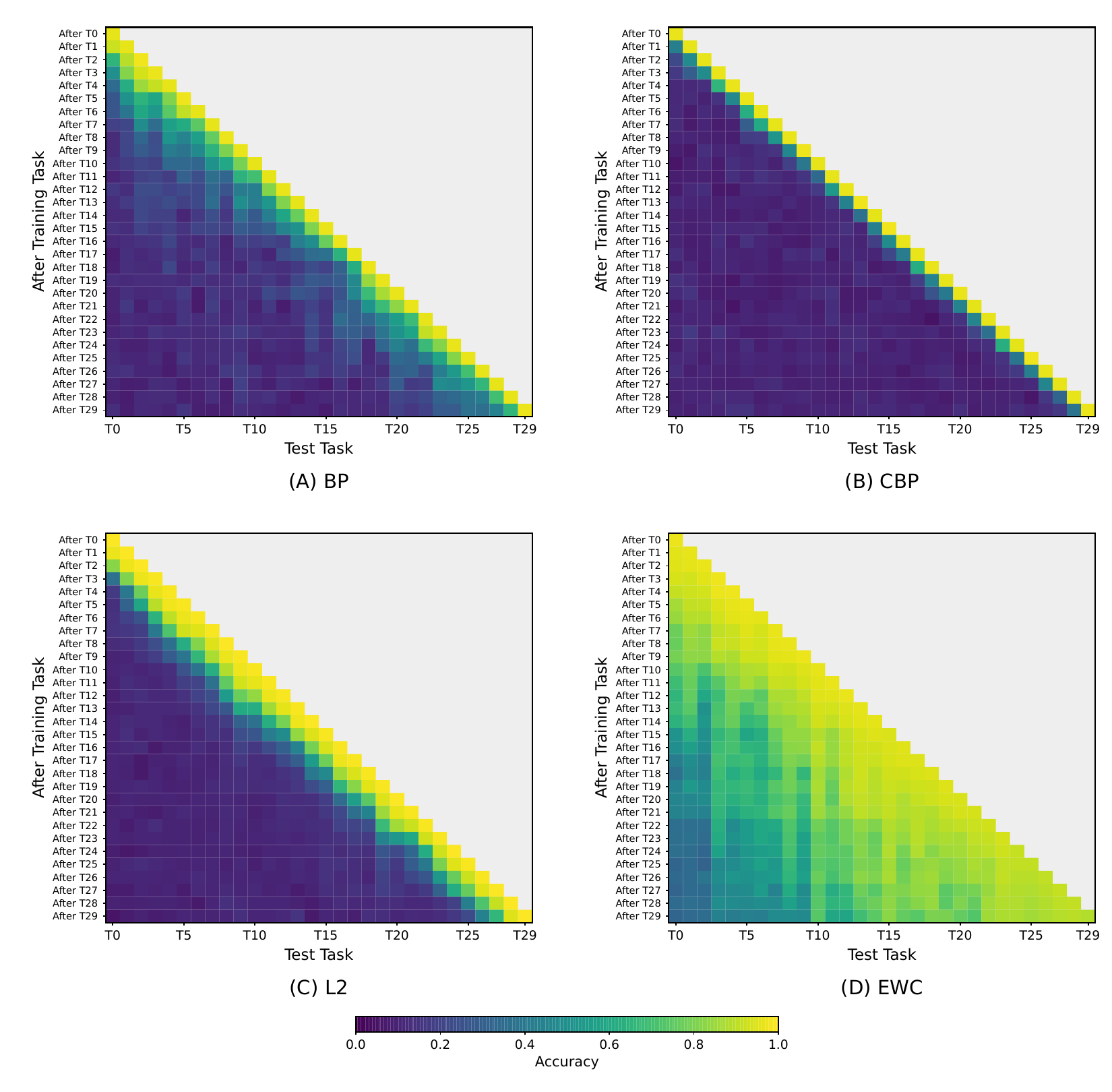}
\caption{The task-by-task test accuracy matrix for four continual learning algorithms. The horizontal axis denotes the index of tested historical tasks, and the vertical axis represents the moment after training the $x$-th task. Specifically, after training each $x$-th task, the current model is tested on all previously learned historical tasks ($0\sim x-1$). The color of each block indicates the classification accuracy of the corresponding task, with the right‑side color bar mapping accuracy values (yellow for high accuracy, purple for low accuracy). Subplots (A)–(D) correspond to BP, CBP, L2, and EWC algorithms, respectively, to intuitively compare forgetting characteristics on historical tasks across different algorithms.}
  \label{fig:accuracy_matrix}
\end{figure}

\subsection{Experimental setup}

All networks employed ReLU activation for hidden layers and softmax activation for the output layer, and were trained with SGD using cross-entropy loss. For the primary benchmark, Permuted MNIST\cite{ref1}, we used a fully connected feed-forward network with three hidden layers of 2000 neurons each. This network processed thirty sequential tasks; each task was generated by applying a fixed random permutation to the 784 input pixels of the original MNIST dataset, preserving the standard 60,000/10,000 train/test split. We adopt the task-incremental learning (Task-IL) protocol: during both training and evaluation, task identity is assumed known, and the model is evaluated on the task-specific output head corresponding to the test sample's task label. Every algorithm—standard BP\cite{ref15}, L2\cite{ref16}, CBP\cite{ref7}, and EWC—was evaluated under identical initialization seeds and a common training protocol (learning rate 0.01 , batch size 1, ten epochs per task), differing only in algorithm-specific regularization or update rules. The architecture and task protocol for the ResNet-18/TinyImageNet validation are detailed in SI Appendix S2.2. All remaining hyperparameter settings, including regularization strengths, task-specific training schedules, and algorithm-specific configuration details for L2, CBP, and EWC, are provided in the Supplementary Information .

\subsection{Network Capacity as the Structural Feasibility Premise}
A foundational yet often overlooked premise in continual-learning research is that any neural architecture of fixed parameter scale P possesses an inherent capacity ceiling. For a given task distribution, there exists an upper bound on the number of distinct tasks, beyond which the network can no longer form internally separable and classifiable representations, regardless of the training protocol employed. This ceiling is not an artifact of optimization failure or hyper-parameter mismatch; rather, it reflects a structural constraint: the representational volume afforded by P parameters is finite.

The significance of formalizing network capacity lies in its role as a hard feasibility filter for the tri-protocol framework. Among the three regimes, joint training is the most structurally privileged: it enjoys simultaneous, unrestricted access to all task data, thereby sidestepping sequential interference entirely. Consequently, joint training defines the best-case representational upper bound achievable under scale P. If the predetermined task count T in an experimental benchmark already exceeds the capacity limit—such that even this omniscient joint-training oracle suffers collapsed accuracy or severely degraded NC metrics—then performance failures in the independent or continual regimes cannot be meaningfully attributed to algorithmic forgetting or insufficient regularization. Instead, the system has entered a regime of capacity saturation, where the very feasibility of learning is compromised.

Therefore, before any diagnosis of representational drift or geometric degradation under sequential learning, it is essential to establish that the evaluation protocol operates within the network’s admissible capacity range. Only when joint training itself yields viable geometry—satisfactory accuracy alongside well-formed $\text{NC}_1$–$\text{NC}_4$ signatures—does the independent boundary represent a true isolation cost, and does the continual boundary reveal genuine algorithmic deficiency. In this framework, capacity does not merely benchmark performance; it demarcates the boundary between evaluable algorithmic behavior and structurally infeasible demands.

\subsection{Task completion criterion}
Across all training protocols—joint, independent, and continual—we adopt a uniform criterion to determine when a task is considered learned. For the current task, training proceeds epoch by epoch. After each epoch, we evaluate the network's accuracy on the current task. If the accuracy reaches a preset threshold $\eta$, training for the current task terminates immediately. Otherwise, training continues for the next epoch. This process repeats until either the accuracy threshold $\eta$ is met or the epoch count reaches a maximum limit $\tau$. The same ($\eta$, $\tau$) rule governs every task in joint training, independent training, and continual training, ensuring a consistent and comparable stopping condition across the tri-protocol boundary.

\subsection{Algorithms}

In this study, we systematically compare five representative learning strategies that tackle the forgetting–plasticity dilemma through distinct mechanisms:
\begin{itemize}
    \item \textbf{Standard Backpropagation (Standard BP)\cite{ref15}}: The vanilla baseline. It minimizes the standard cross-entropy loss without any continual-learning-specific mechanism:
$$
\mathcal{L}_{\text{BP}} = -\frac{1}{N} \sum_{n=1}^{N} \sum_{c=1}^{C} y_{n,c} \log(p_{n,c})
$$
where $p_{n,c}$ denotes the softmax probability of sample $n$ belonging to class $c$, and $y_{n,c}$ is the corresponding one-hot label. Under sequential training, this represents the default behavior of deep networks, providing no explicit protection against catastrophic forgetting.

    \item \textbf{L2-regularized Backpropagation (L2)\cite{ref16}}: A parameter-constraint method that mitigates forgetting by augmenting the objective with an $\text{L}_2$ penalty on parameter magnitudes:
$$
\mathcal{L}_{\text{L2}} = \mathcal{L}_{\text{BP}} + \frac{\lambda}{2} \|\theta\|_2^2
$$
The additional term $\frac{\lambda}{2} \|\theta\|_2^2$ drives weights toward the origin, restricting the magnitude of subsequent parameter updates and implicitly confining the solution space to a neighborhood that favors cross-task stability.

    \item \textbf{Continual Backpropagation (CBP)\cite{ref7}}:Designed to enhance plasticity by actively promoting parameter turnover. CBP retains the standard objective $\mathcal{L}_{\text{BP}}$, but interleaves each gradient step with a stochastic replacement operator:
$$
\theta_{t+1} = \mathcal{R}_s \left( \theta_t - \eta \nabla_\theta \mathcal{L}_{\text{BP}} \right)
$$
Here $\mathcal{R}_s$ identifies the fraction $s$ of parameters with the lowest utility—typically measured via activation-gradient products—and reinitializes them. By continually injecting randomness into dormant weights, CBP preserves parametric mobility for incoming tasks while attempting to retain old knowledge.

    \item \textbf{Elastic Weight Consolidation (EWC)\cite{ref5}}: A prominent anti-forgetting strategy that selectively protects parameters deemed important for previously learned tasks. It augments the current-task loss with a quadratic penalty anchored at the old-task optimum:
$$
\mathcal{L}_{\text{EWC}} = \mathcal{L}_{\text{BP}} (\theta; D_{\text{new}}) + \frac{\lambda}{2} \sum_{i} F_i \left( \theta_i - \theta_i^* \right)^2
$$
where $\theta^*$ denotes the parameters after training on previous tasks, and $F_i$ is the diagonal Fisher information approximating the importance of parameter $i$ for past tasks. This penalty effectively creates a parabolic energy well around critical parameters, prioritizing stability through knowledge consolidation.

    \item \textbf{Predictive Distillation (PredKD)\cite{ref19,ref21}}: A function-space regularization strategy that constrains the predictive response instead of the parameters. At each task boundary, the converged model of the previous task is frozen as a teacher, and the current-task loss is augmented with a temperature-scaled divergence:
\[
\begin{aligned}
\mathcal{L}_{\text{PredKD}}
&= \mathcal{L}_{\text{BP}} (\theta; D_{\text{new}}) \\
&\quad + \alpha T^2 \, \mathrm{KL}\!\left(
\mathrm{softmax}\!\left(\frac{z_t}{T}\right)
\Big\|
\mathrm{softmax}\!\left(\frac{z_\theta}{T}\right)
\right)
\end{aligned}
\]
where $z_\theta$ and $z_t$ denote the logits of the student and of the frozen teacher, $T$ is the temperature that softens both distributions, and $\alpha$ weights the distillation term. Penalising the realised function rather than parameter coordinates keeps the backbone free to adapt while preserving the predictive structure of earlier tasks.
\end{itemize}

By evaluating these diverse strategies within our boundary-anchored benchmark, we provide a direct comparison of how different approaches to the forgetting–plasticity problem manifest in terms of internal geometric reorganization.

\subsection{Neural collapse metrics}

For each task, let $f_\theta$ denote the feature extractor and let $h_{k,i}=f_\theta(x_{k,i})\in\mathbb{R}^d$ be the penultimate-layer feature of the $i$-th sample belonging to class $k$ ($k=1,\dots,K$). The class mean and global mean are $\mu_k=\frac{1}{N_k}\sum_{i=1}^{N_k}h_{k,i}$ and $\mu_G=\frac{1}{K}\sum_{k=1}^K\mu_k$, respectively. Define centered class means $\tilde\mu_k=\mu_k-\mu_G$ and, analogously, centered classifier weight vectors $\tilde w_k$ obtained by removing the mean of the final-layer weight matrix $W$.

\textbf{$\text{NC}_1$ (within-class variability)\cite{ref10}.}
\[
\text{$\text{NC}_1$}=\frac{\frac{1}{K}\sum_{k=1}^K\frac{1}{N_k}\sum_{i=1}^{N_k}\|h_{k,i}-\mu_k\|_2^2}{\frac{1}{K}\sum_{k=1}^K\|\tilde\mu_k\|_2^2}.
\]
This ratio compares the average within-class covariance to the average between-class covariance. A value approaching zero indicates tight feature collapse around class means.

\textbf{$\text{NC}_2$ (class-mean geometry).}
This criterion is decomposed into two complementary components after centering.

\noindent Equinormality (EN)\cite{ref11}:
\[
\text{EN}=\frac{1}{K}\sum_{k=1}^K\Big|\|\tilde\mu_k\|_2-\frac{1}{K}\sum_{j=1}^K\|\tilde\mu_j\|_2\Big|.
\]

\noindent Equiangularity (EA)\cite{ref11}:
Let $c=-\frac{1}{K-1}$ denote the theoretical cosine of the maximum equiangular separation in a Simplex ETF. Then
\[
\text{EA}=\frac{1}{K(K-1)}\sum_{k=1}^K\sum_{j\neq k}\Big|\frac{\langle\tilde\mu_k,\tilde\mu_j\rangle}{\|\tilde\mu_k\|_2\|\tilde\mu_j\|_2}-c\Big|.
\]
Low EN and EA indicate that class means are approaching the regular simplex configuration.

\textbf{$\text{NC}_3$ (classifier--feature duality)\cite{ref10}.}
\[
\text{$\text{NC}_3$}=\frac{1}{K}\sum_{k=1}^K\Big(1-\frac{\langle\tilde w_k,\tilde\mu_k\rangle}{\|\tilde w_k\|_2\|\tilde\mu_k\|_2}\Big).
\]
$\text{NC}_3$ measures the cosine alignment between the classifier weights and the corresponding centered class means; it approaches zero when the classifier self-duality characteristic of neural collapse is realized.

\textbf{$\text{NC}_4$ (nearest-class-center consistency)\cite{ref10}.}
For a given sample with feature $h$, the nearest-class-center classifier assigns $\hat y=\arg\min_k\|h-\mu_k\|_2$. $\text{NC}_4$ is defined as the fraction of test samples for which the NCC prediction disagrees with the actual network prediction, so that a low $\text{NC}_4$ indicates that the parametric classifier mimics the NCC decision rule.

\section*{Acknowledgment}

This work is supported by the New Generation Artificial Intelligence-National Science and Technology Major Project (2026ZD0126104) of Ministry of Science and Technology of the People's Republic of China (MOST).


\end{document}